\PassOptionsToPackage{table}{xcolor}
\documentclass[11pt]{article}

\usepackage{acl}
\usepackage{times}
\usepackage{latexsym}
\usepackage{booktabs}
\usepackage{multirow}
\usepackage{amsmath}
\usepackage{makecell}
\usepackage{graphicx}
\usepackage{booktabs}
\usepackage{subcaption}

\newcommand{\bootci}[3]{%
  \shortstack[c]{%
    $#1$\\[-1pt]
    {\scriptsize $[#2,#3]$}%
  }%
}
\usepackage{tabularx}
\newcolumntype{R}{>{\raggedleft\arraybackslash}X}
\newcolumntype{C}{>{\centering\arraybackslash}X}

\usepackage[T1]{fontenc}

\usepackage[utf8]{inputenc}

\usepackage{microtype}

\usepackage{inconsolata}

\usepackage{graphicx}
\usepackage{pifont}

\usepackage{tcolorbox}
\tcbuselibrary{skins,breakable,listings}
\newtcblisting{promptbox}[1][]{%
  enhanced, breakable,
  colback=gray!5, colframe=gray!55,
  boxrule=0.4pt, arc=1.2pt,
  left=6pt, right=6pt, top=4pt, bottom=4pt,
  listing only,
  listing options={
    basicstyle=\ttfamily\small,
    breaklines=true,
    breakatwhitespace=true,
    columns=fullflexible,
    keepspaces=true,
    showstringspaces=false,
    aboveskip=0pt, belowskip=0pt,
  },
  #1
}

\title{Emotion is not External Knowledge:\\ Internal Attention Signals for Affective State Reasoning}

\author{
  Wenqiu Tang\textsuperscript{1}\quad
  Zhen Wan\textsuperscript{2} \quad
  Yihui Wang\textsuperscript{1} \\
  \textbf{Takahiro Komamizu}\textsuperscript{1} \quad
  \textbf{Ichiro Ide}\textsuperscript{1} \\
  \textsuperscript{1}Nagoya University, Nagoya, Aichi, Japan \\
  \textsuperscript{2}National Institute of Informatics, Tokyo, Japan \\
  \texttt{\{tangw, wangyi\}@cs.is.i.nagoya-u.ac.jp} \\
  \texttt{\{taka-coma@acm.org, ide@i.nagoya-u.ac.jp, zhenwan.nlp@gmail.com\}}
}

\setcitestyle{citesep={,}}

\begin{document}
\maketitle
\begin{abstract}
Large Language Models (LLMs) increasingly operate in multi-turn interactions that require maintaining latent conversational states, not only explicit facts. We study affective state as an implicit state inferred from sparse contextual cues that remain relevant after topic shifts, and ask whether LLMs expose internal attention signals for routing such evidence at inference time.
Prior attention interventions rely on head profiles derived from labeled examples across multiple profiling tasks and externally specified target spans, limiting transfer across affective-evidence distributions. We propose Self-targeted Internal Attention Boosting (SIAB), which localizes sparse affective heads across transformer layers in each model via causal localization (in detail activation patching) on synthetic minimal pairs and boosts the content tokens these heads already attend to. Both edited heads and target tokens are derived from the model's own behavior, enabling the same head set to be reused across tasks. We evaluate SIAB on the ECC and GoEmotions benchmarks across three LLMs and multiple prompting settings. Relative to the corresponding settings without SIAB, macro-F1 improves by up to 7.67 points on ECC and 3.28 points on GoEmotions. Combining SIAB with PASTA further raises ECC macro-F1 by 12.51 points over the Gemma 3 zero-shot baseline.
\end{abstract}

\section{Introduction}
\label{sec:intro}

\begin{figure}[t]
    \centering
    \includegraphics[
        width=\linewidth,
        trim=0 2.0cm 0 0,
        clip
    ]{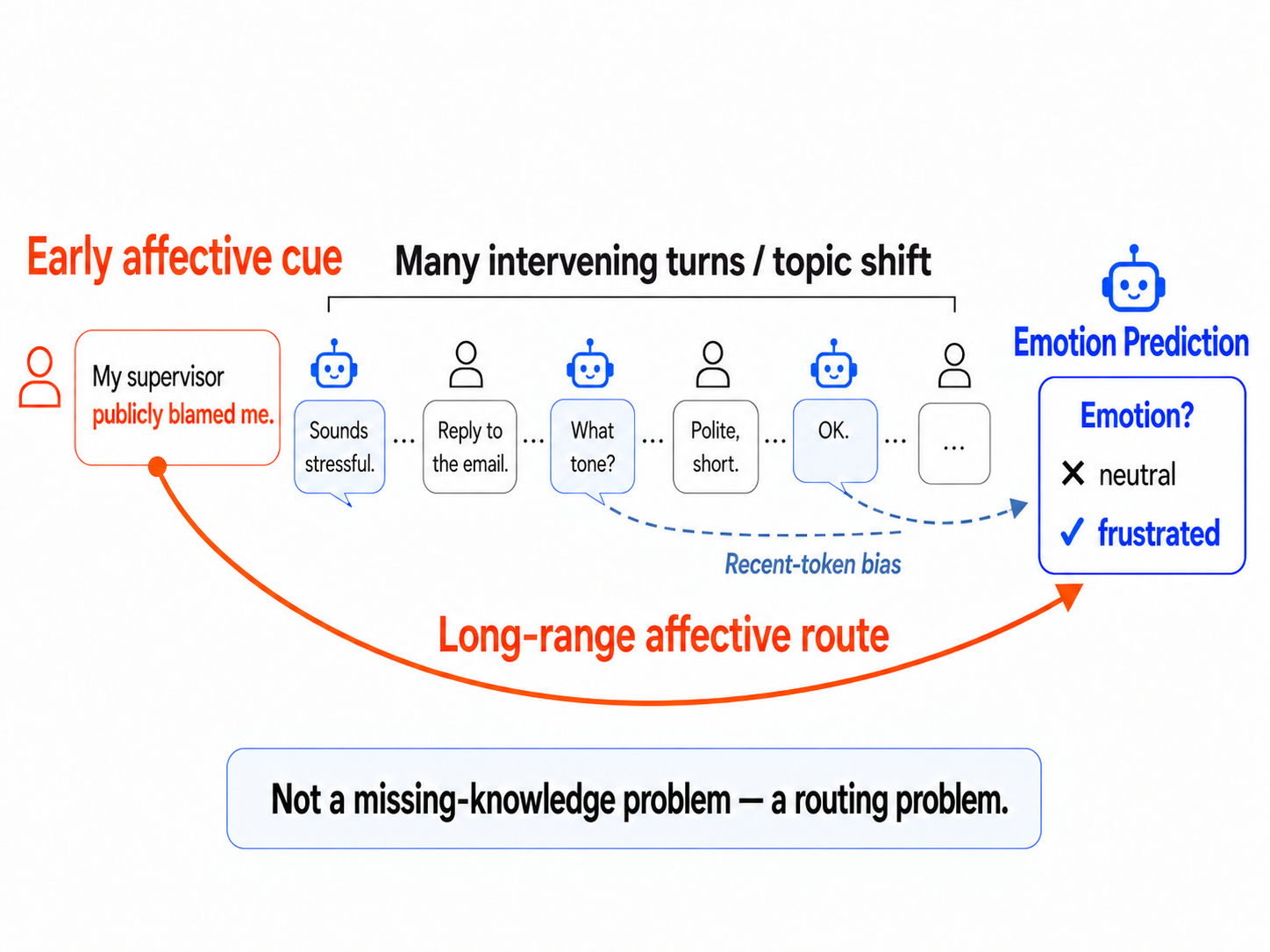}
    \caption{
Long dialogues can bury early affective cues, turning affective prediction into a routing problem.
}
    \label{fig:motivation}
\end{figure}

Conversation is inherently stateful; What a user means at any point
depends on what came before. As Large Language Models (LLMs) are
increasingly deployed in such multi-turn interaction, successful
responses must not only retain explicit factual context but also
maintain latent conversational states, such as a user's frustration,
uncertainty, or trust~\citep{fazzi2025emotions,folstad2024breakdown}.
Affective state is a challenging case. It is rarely given
as an explicit variable. Instead, it must be inferred from sparse
contextual cues whose relevance can persist across topic shifts. As
illustrated in Figure~\ref{fig:motivation}, a model may therefore need
to use an earlier affective cue when making a later prediction.

Unlike knowledge-intensive failures, where retrieval or tool use can
supply a missing fact, affective continuity does not require adding new
information. Since the relevant evidence is already in the dialogue, the
difficulty is whether the model routes it to the final decision point.
Emotion recognition therefore provides a controlled proxy for studying
this routing problem at inference time.

Inference-time attention intervention is a natural way to probe this
problem, since it reroutes existing contextual evidence during the
forward pass, without parameter updates or newly supplied facts. Existing
attention intervention methods, however, rely on external control
signals. Post-hoc Attention STeering Approach (PASTA)
\citep{zhang2024pasta} profiles which heads to edit from labeled
in-domain data and steers attention toward externally specified spans.
Later variants reduce but do not remove this reliance on external
signals. These methods raise two questions. RQ1: Can internal attention signals
route affective evidence without external spans? and RQ2: Are
internal and external signals redundant or complementary?

We address these questions with Self-targeted Internal Attention Boosting
(SIAB), an attention intervention proposed for affective state reasoning. It first localizes a sparse
set of emotion-sensitive attention heads in different Transformer layers through causal localization via activation patching. It then uses these heads themselves
to select target tokens. At inference time, SIAB adds a scalar bias to
the pre-softmax attention logits of the localized heads, boosting the
content tokens they already attend to. Both the edited heads and the
target tokens thus derive from the model's own behavior.

We evaluate on two Emotion Recognition in Conversation (ERC) settings with complementary distributions of affective evidence. In the Emotion-Cause Conversation benchmark (ECC; \citealp{he2025ecc}), cues are dispersed across earlier turns and must survive later topic shifts. Meanwhile, in the GoEmotions benchmark~\citep{demszky2020goemotions}, a single dense cue must instead be disambiguated among many fine-grained categories. The same head set is reused across both.

Since our goal is not only to improve recognition but to understand the
internal signal we exploit, we analyze when and why the localized heads
work. The gains come from a sparse set of heads; A few heads are sufficient,
and they attend more to affective words than random heads from early,
middle, and late layers. Whether SIAB succeeds further depends on logit
geometry and on whether the boosted tokens align with the final label
decision. Finally, the effects of SIAB and PASTA complement each other depending on their layer-band proximity.

Our contributions are threefold: (i) We propose SIAB, a self-targeted
inference-time attention intervention that requires no parameter update
or annotated span and composes with prompting and PASTA-style external
supervision, (ii) We show consistent gains across models, datasets, and
prompting settings, and (iii) We characterize its mechanism and boundary
conditions.

\begin{figure*}[!t]
\centering
\includegraphics[width=\textwidth]{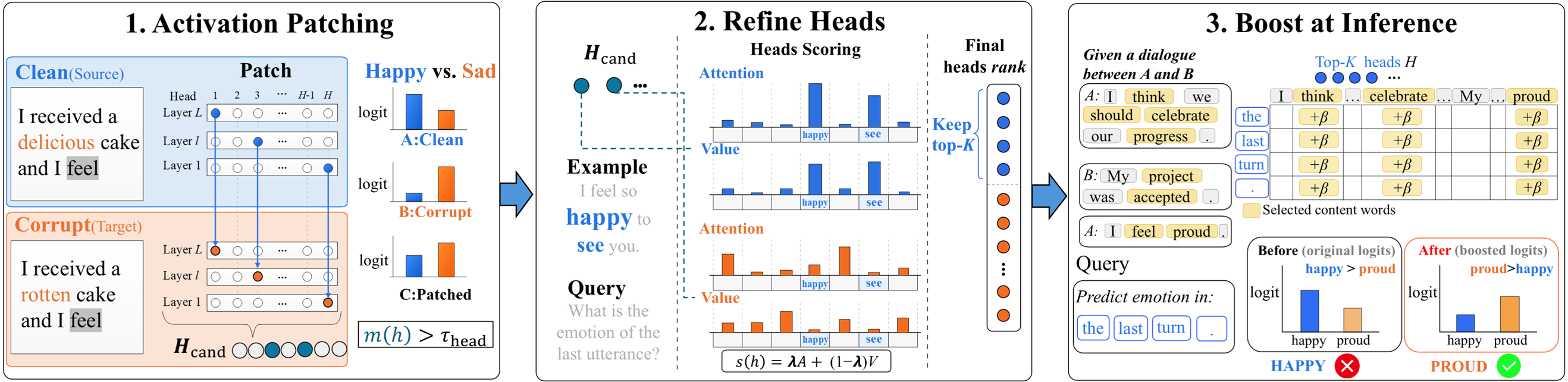}
\caption{Overview of the proposed Self-targeted Internal Attention Boosting (\textsc{SIAB}).
Causal localization via activation patching selects sparse emotion-sensitive heads (4--20 of hundreds).
SIAB adds a pre-softmax bias to the content tokens these heads already attend to
at final query positions ---without training, parameter update, or external span.}
\label{fig:overview}
\end{figure*}
\section{Related Work}
\label{sec:related}
Existing attention-level methods externalize control to varying degrees,
relying on heads selected by testing each head’s steering effect on development examples with target tokens that are user-annotated
\citep{zhang2024pasta}, heuristically extracted
\citep{zhang2024autopasta,yu2024act}, or applied at the head output rather
than its attention \citep{li2023iti}. Residual-stream steering operates at
a coarser granularity
\citep{turner2024steering,rimsky2024caa,zou2023representation}. These
methods assume that the relevant evidence can be externally identified, an
assumption less natural for affective state, whose cues spread across
turns, often lack lexical salience, and shift as the conversation unfolds.
We instead derive both signals internally ---heads from activation patching,
target tokens from each head's own attention--- and study how this internal
signal composes with external ones. 

Mechanistic interpretability provides tools for discovering internal
structures that mediate model behavior. Activation patching
\citep{wang2023ioi,goldowsky2023localizing} has been used to localize
concrete circuit components, such as name-mover and S-inhibition heads,
through clean--corrupted activation contrasts, while direct logit attribution
\citep{nostalgebraist2020logit} links intermediate contributions to
final-token logits. These tools fit our goal of identifying internal signals
for affective-evidence routing. Rather than specifying emotion-bearing spans
externally, we localize the attention heads that internally route affective
contrast.

Emotion recognition in conversation naturally probes affective state
across dialogue turns, yet existing mechanistic work operates at a
coarser granularity. It studies sentiment
directions~\citep{tigges2024sentiment}, mid-layer multi-head
self-attention units~\citep{tak2025mechanistic}, or residual-stream
circuits for emotion \emph{generation}~\citep{wang2025emotion}, leaving
head-level recognition unaddressed. Other work improves accuracy through
retrieval, speaker modeling, or curriculum
learning~\citep{lei2023instructerc,xue2024bioserc,fu2025laerc,li2025prcemo},
but treats the model as a black box and ignores its internal attention
signal.

\section{Method}

We introduce \textsc{SIAB}, an inference-time attention intervention with two components: Head-set construction via Activation Patching and
heads refinement introduced in \S\ref{sec:head-selection} and \S\ref{sec:refinement}, respectively, and Self-targeted pre-softmax logit boosting introduced in \S\ref{sec:intervention}. Both components derive their control signal from the model's own internal behavior. Figure~\ref{fig:overview} gives an overview.

We frame emotion prediction as generative label scoring. Given an input
context $X$ and a candidate label set $\mathcal{Y}$, we score each label
using its answer-position verbalizer logit, rather than training a
task-specific classification head. 






\subsection{Head Selection via Activation Patching}
\label{sec:head-selection}
 
Activation patching localizes the attention heads in different Transformer layers that causally affect
the original prediction using prompt pairs that differ only in the feature
driving that prediction.
In our case, the difference in affective state is driven by the
affective cues, so we build clean--corrupted pairs
$(X^{\mathrm{clean}}, X^{\mathrm{corr}})$, e.g.,
$X^{\mathrm{clean}}$\,=\,\emph{``I received a delicious cake. I feel''}
and $X^{\mathrm{corr}}$\,=\,\emph{``I received a rotten cake. I feel''}~(Appendix~\ref{app:pair_construction} shows the detail of pairs construction).
Two properties make these pairs suitable for causal attribution.
First, every prompt ends with \texttt{I feel}, so the model is induced to predict an emotion word at the final position, where we read off the effect.
Second, the two prompts share the same syntactic structure and differ only in the affective cue, so any difference in the final-position emotion prediction can be attributed to that cue.

We run the model on $X^{\mathrm{corr}}$ as a baseline.
Then, for one head $h$ at a time, we replace \emph{only} that head's final-token output with the output it produced on the $X^{\mathrm{clean}}$ and complete the forward pass. Everything else stays at its corrupted value.
If $h$ encodes the cue, restoring its clean output pulls the emotion prediction back toward the clean answer. Otherwise, the model's prediction should remain close to the
corrupted-run baseline.
We quantify this as the normalized logit-difference recovery formulated as:
\begin{align}
m(h) &= \frac{\mathcal{L}(\mathrm{patch}_h) - \mathcal{L}(X^{\mathrm{corr}})}{\mathcal{L}(X^{\mathrm{clean}}) - \mathcal{L}(X^{\mathrm{corr}})}, \\
\mathcal{L}(\cdot) &= \mathrm{logit}(e_{\mathrm{clean}}) - \mathrm{logit}(e_{\mathrm{corr}}). \nonumber
\end{align}
Here, $e_{\mathrm{clean}}$ and $e_{\mathrm{corr}}$ are the top-ranked
final-position emotion words on $X^{\mathrm{clean}}$ and
$X^{\mathrm{corr}}$, respectively, and $\mathrm{patch}_h$ denotes the
patched run described above.
Thus $m(h)=1$ when the swap fully restores the clean prediction and $m(h)=0$ when it has no effect, following~\citet{nostalgebraist2020logit} and~\citet{tigges2023linear}.
We apply the test to every head independently and retain those with $m(h) > \tau_{\mathrm{head}}$, and their union forms the candidate set $\mathcal{H}_{\mathrm{cand}}$.
\subsection{Label-free Refinement on Dialogues}
\label{sec:refinement}
Having obtained the candidate heads $\mathcal{H}_{\mathrm{cand}}$, we do
not directly treat all of them as suitable intervention targets. A head
may obtain a high patching score $m(h)$ through
generic mechanisms, such as positional tracking or induction-like
copying~\citep{olsson2022induction}, rather than by routing affective
cues. In natural dialogues, affective
evidence is usually carried by lexical content. We therefore use a lightweight content-token filter that removes punctuation, special tokens, pronouns, interrogatives, and prepositions, while retaining negators and intensifiers. We then refine the candidates on dialogues by measuring
their attention and value mass on content-word positions before the
emotion prediction, using value mass because attention alone does not
determine the information written to the residual stream~\citep{kobayashi2020attention}.

For this refinement stage, we use 200 dialogues sampled from the DailyDialog dataset~\citep{li-etal-2017-dailydialog}. For each dialogue, we append a probe question, e.g., \emph{``What is the emotion of the last utterance?''}, and inspect each candidate head immediately before the model generates the answer. Although this dataset provides emotion annotations, we use only the dialogue text during this stage and do not access any gold emotion label.

For an arbitrary token position $u$, let $\alpha_{h,q,u}$ be the
attention from query position $q\in\mathcal{Q}$ to $u$ on head $h$, and
let $v_{h,q,u}$ be the L2 norm of the corresponding value contribution.
Because dialogue turns have different lengths, we normalize both
quantities within the conversational turn as:
\begin{equation}
\begin{aligned}
\tilde\alpha_{h,q,u} &=
\frac{\alpha_{h,q,u}}
{\sum_{u'\in\mathrm{turn}(u)}\alpha_{h,q,u'}}, \\
\tilde v_{h,q,u} &=
\frac{v_{h,q,u}}
{\sum_{u'\in\mathrm{turn}(u)}v_{h,q,u'}},
\end{aligned}
\end{equation}
where $\mathrm{turn}(u)$ denotes all token positions in the same
conversational turn as $u$, including both content and non-content
tokens. Thus, $\tilde\alpha_{h,q,u}$ and $\tilde v_{h,q,u}$ measure the
fraction of a head's attention and value mass within that turn assigned
to $u$.

Let $\mathcal{C}(r)$ denote the content-word positions in turn $r$. The
candidate score is defined as:
\begin{equation}
s_{\mathrm{head}}(h)=\lambda A_h+(1-\lambda)V_h,
\label{eq:head-score}
\end{equation}
where $A_h$ and $V_h$ are computed by summing the turn-normalized masses
over $u\in\mathcal{C}(r)$ and then averaging over turns $r$, query
positions $q\in\mathcal{Q}$, and refinement dialogues. This yields the fraction of each head's turn-level mass assigned to
substantive dialogue content. The final head set $\mathcal{H}$ is the top-$K$ heads by
$s_{\mathrm{head}}$. Per-model values of $K$ and the resulting head
indices are reported in Appendix~\ref{app:refinement}.
\subsection{Attention Logit Boosting}
\label{sec:intervention}

Crucially, the intervened tokens are not externally annotated, but selected
from each chosen head's own attention distribution at the final query
positions. At inference time, for each
$(h,q)\in\mathcal{H}\times\mathcal{Q}$, let $\mathcal{T}_r(h,q)$ denote
the top-$r$ fraction of content-word tokens ranked by attention score
$\alpha_{h,q,t}$. We add a scalar bias $\beta$ to the pre-softmax
attention logits $\ell_{h,q,t}$ of these tokens as:
\begin{equation}
\ell'_{h,q,t}
=
\ell_{h,q,t}
+
\beta \mathbf{1}[t\in\mathcal{T}_r(h,q)].
\label{eq:attention-bias}
\end{equation}
The attention distribution is then computed by applying softmax to the
modified logits $\ell'_{h,q,t}$.

The head set $\mathcal{H}$ is fixed once and reused across both datasets
without re-localization. Also the bias magnitude $\beta$ is calibrated per
model and then held fixed across both datasets. Hyperparameter
values and the calibration protocol are reported in
Appendix~\ref{app:hparams}.
\section{Experimental Evaluation}
\label{sec:setup}
\subsection{Datasets}
\label{sec:datasets}
We evaluate on two benchmarks that stress different cue regimes.
\textbf{ECC}~\citep{he2025ecc} labels multi-turn dialogues
(max.\ $831$ tokens) with $10$ fine-grained emotion classes, where
affective evidence can appear early and must remain available across many
turns. \textbf{GoEmotions}~\citep{demszky2020goemotions} labels short Reddit
comments\footnote{\url{https://www.reddit.com/} (Accessed: Mar. 19, 2026)} (max.\ $52$ tokens) with $27$ multi-label emotion categories plus
a Neutral class, where cues are local and information-dense but must be
separated among many fine-grained labels. We exclude samples labeled \emph{solely} as Neutral, since
they contain no non-neutral affective cue for cue-targeting heads to act on.
For multi-label samples, we retain the non-Neutral labels. This leaves
$3{,}821$ test examples for evaluation. Neutral results are
reported in Appendix~\ref{app:neutral}.

\subsection{Evaluation Metrics}
We report macro-F1 and Accuracy. For the ECC benchmark, we take the 
argmax over 10 labels. For the GoEmotions benchmark, we follow the multi-label setting of 
\citet{demszky2020goemotions}. The
temperature $T$ and threshold $t$ are tuned on the GoEmotions dev split (The specific hyperparameter values are provided in Appendix~\ref{app:hparams}). and Top-$1$ accuracy counts the argmax label as 
correct if it belongs to the gold label set.

We additionally quantify finite-sample uncertainty using paired bootstrap
resampling ($B=10{,}000$). Full pointwise 95\% percentile confidence
intervals and implementation details are reported in
Appendix~\ref{app:bootstrap}.
\subsection{Models}
We evaluate three open, small-to-mid-sized instruction-tuned LLMs from different model families: Gemma3-4B~\citep{gemma2025technical}, Qwen3-4B~\citep{yang2025qwen3}, and LLaMA3-8B~\citep{grattafiori2024llama3}. 

\begin{table*}[!t]
\centering
\caption{Results of two different metrics on the ECC~\citep{he2025ecc} and  GoEmotions~\citep{demszky2020goemotions} benchmarks.
\textit{ZS} / \textit{Mid} / \textit{High} denote zero- / few-shot prompts
(15/27 for GoEmotions, 5/10 for ECC).
\textit{PASTA-\{ECC,GoE\}} use heads selected on the named dataset,
\textit{PASTA-$\cap$} keeps the shared top heads from these two profiles.
\textit{Rand.} is layer-matched random.
Scores are $\times100$, $\Delta$ is vs.\ ZS, best per model in \textbf{bold}.}
\label{tab:merged-results}

\small
\setlength{\tabcolsep}{4pt}

\newcommand{\scorecell}[2]{%
  \makebox[4.6em][c]{%
    \makebox[1.30em][r]{#1}%
    \makebox[0.22em][c]{.}%
    \makebox[0.90em][l]{#2}%
  }%
}

\newcommand{\bscorecell}[2]{%
  \makebox[4.6em][c]{%
    \makebox[1.30em][r]{\textbf{#1}}%
    \makebox[0.22em][c]{\textbf{.}}%
    \makebox[0.90em][l]{\textbf{#2}}%
  }%
}

\newcommand{\scorecellpm}[3]{%
  \makebox[4.6em][c]{%
    \makebox[1.30em][r]{#1}%
    \makebox[0.22em][c]{.}%
    \makebox[0.90em][l]{#2}%
    \rlap{\hspace{0.10em}\ensuremath{\scriptscriptstyle\pm#3}}%
  }%
}

\newcommand{\deltacell}[3]{%
  \makebox[3.8em][c]{%
    \makebox[1.70em][r]{%
      \ifx#1-%
        \scalebox{0.85}[1]{$-$}#2%
      \else
        #1#2%
      \fi
    }%
    \makebox[0.22em][c]{.}%
    \makebox[0.90em][l]{#3}%
  }%
}

\newcommand{\bdeltacell}[3]{%
  \makebox[3.8em][c]{%
    \makebox[1.70em][r]{%
      \ifx#1-%
        \scalebox{0.85}[1]{$\mathbf{-}$}\textbf{#2}%
      \else
        \textbf{#1#2}%
      \fi
    }%
    \makebox[0.22em][c]{\textbf{.}}%
    \makebox[0.90em][l]{\textbf{#3}}%
  }%
}

\newcommand{\dashcell}{\multicolumn{1}{c}{\hspace{0.5em}---}}

\newcommand{\modelcell}[2]{%
  \rotatebox{90}{%
    \begin{tabular}{@{}c@{}}
      \textbf{#1}\\[-1pt]
      #2
    \end{tabular}%
  }%
}

\begin{tabular}{llcccccccc}
\toprule
& &
\multicolumn{4}{c}{\textbf{ECC} benchmark} &
\multicolumn{4}{c}{\textbf{GoEmotions} benchmark} \\
\cmidrule(lr){3-6} \cmidrule(lr){7-10}

\textbf{Model} & \textbf{Config}
& \textbf{ Acc.}
& \textbf{ Macro-F1}
& \textbf{ $\Delta_{\text{Acc}}$}
& \textbf{ $\Delta_{\text{F1}}$}
& \textbf{ Macro-F1}
& \textbf{ Top-1}
& \textbf{ $\Delta_{\text{F1}}$}
& \textbf{ $\Delta_{\text{Top1}}$} \\

\midrule

\multirow{11}{*}{%
  \modelcell{Gemma3}{\citep{gemma2025technical}}%
}

& ZS
& \scorecell{50}{29}
& \scorecell{52}{26}
& \dashcell
& \dashcell
& \scorecell{27}{17}
& \scorecell{34}{37}
& \dashcell
& \dashcell \\

& ZS+SIAB
& \scorecell{59}{62}
& \scorecell{59}{93}
& \deltacell{+}{9}{32}
& \deltacell{+}{7}{67}
& \scorecell{30}{45}
& \bscorecell{38}{15}
& \deltacell{+}{3}{28}
& \deltacell{+}{3}{78} \\

& Mid
& \scorecell{56}{48}
& \scorecell{55}{51}
& \deltacell{+}{6}{19}
& \deltacell{+}{3}{25}
& \scorecell{30}{55}
& \scorecell{33}{45}
& \deltacell{+}{3}{38}
& \deltacell{-}{0}{92} \\

& Mid+SIAB
& \scorecell{59}{90}
& \scorecell{60}{15}
& \deltacell{+}{9}{61}
& \deltacell{+}{7}{89}
& \scorecell{32}{65}
& \scorecell{36}{68}
& \deltacell{+}{5}{47}
& \deltacell{+}{2}{30} \\

& High
& \scorecell{60}{66}
& \scorecell{58}{27}
& \deltacell{+}{10}{37}
& \deltacell{+}{6}{01}
& \scorecell{28}{72}
& \scorecell{32}{24}
& \deltacell{+}{1}{55}
& \deltacell{-}{2}{14} \\

& High+SIAB
& \scorecell{63}{69}
& \scorecell{62}{52}
& \deltacell{+}{13}{40}
& \deltacell{+}{10}{27}
& \scorecell{31}{49}
& \scorecell{36}{19}
& \deltacell{+}{4}{32}
& \deltacell{+}{1}{82} \\

\cmidrule(lr){2-10}

& PASTA-ECC
& \scorecell{59}{43}
& \scorecell{60}{37}
& \deltacell{+}{9}{14}
& \deltacell{+}{8}{11}
& \scorecell{26}{86}
& \scorecell{32}{89}
& \deltacell{-}{0}{31}
& \deltacell{-}{1}{48} \\

& PASTA-GoE
& \scorecell{52}{11}
& \scorecell{52}{99}
& \deltacell{+}{1}{82}
& \deltacell{+}{0}{73}
& \scorecell{31}{81}
& \scorecell{34}{37}
& \deltacell{+}{4}{64}
& \deltacell{+}{0}{00} \\

& PASTA-$\cap$
& \scorecell{59}{12}
& \scorecell{60}{02}
& \deltacell{+}{8}{83}
& \deltacell{+}{7}{76}
& \scorecell{29}{19}
& \scorecell{34}{65}
& \deltacell{+}{2}{02}
& \deltacell{+}{0}{28} \\

& PASTA+SIAB
& \bscorecell{64}{13}
& \bscorecell{64}{77}
& \bdeltacell{+}{13}{84}
& \bdeltacell{+}{12}{51}
& \bscorecell{33}{75}
& \scorecell{38}{01}
& \bdeltacell{+}{6}{58}
& \deltacell{+}{3}{64} \\

\cmidrule(lr){2-10}

& Random
& \scorecellpm{41}{89}{3.70}
& \scorecellpm{45}{60}{4.10}
& \deltacell{-}{8}{40}
& \deltacell{-}{6}{66}
& \scorecellpm{26}{01}{1.80}
& \scorecellpm{32}{13}{2.90}
& \deltacell{-}{1}{16}
& \deltacell{-}{2}{24} \\

\midrule

\multirow{11}{*}{%
  \modelcell{Qwen3}{\citep{yang2025qwen3}}%
}

& ZS
& \scorecell{52}{73}
& \scorecell{55}{25}
& \dashcell
& \dashcell
& \scorecell{35}{51}
& \scorecell{39}{94}
& \dashcell
& \dashcell \\

& ZS+SIAB
& \scorecell{56}{03}
& \scorecell{58}{04}
& \deltacell{+}{3}{30}
& \deltacell{+}{2}{79}
& \scorecell{37}{77}
& \scorecell{41}{31}
& \deltacell{+}{2}{26}
& \deltacell{+}{1}{37} \\

& Mid
& \scorecell{55}{63}
& \scorecell{58}{02}
& \deltacell{+}{2}{90}
& \deltacell{+}{2}{78}
& \scorecell{35}{56}
& \scorecell{40}{38}
& \deltacell{+}{0}{05}
& \deltacell{+}{0}{44} \\

& Mid+SIAB
& \scorecell{57}{71}
& \scorecell{59}{91}
& \deltacell{+}{4}{99}
& \deltacell{+}{4}{66}
& \bscorecell{38}{31}
& \scorecell{42}{69}
& \bdeltacell{+}{2}{79}
& \deltacell{+}{2}{75} \\

& High
& \scorecell{58}{71}
& \scorecell{60}{67}
& \deltacell{+}{5}{98}
& \deltacell{+}{5}{42}
& \scorecell{35}{99}
& \scorecell{43}{25}
& \deltacell{+}{0}{48}
& \deltacell{+}{3}{31} \\

& High+SIAB
& \scorecell{60}{36}
& \scorecell{61}{88}
& \deltacell{+}{7}{63}
& \deltacell{+}{6}{63}
& \scorecell{38}{23}
& \bscorecell{44}{26}
& \deltacell{+}{2}{71}
& \bdeltacell{+}{4}{32} \\

\cmidrule(lr){2-10}

& PASTA-ECC
& \scorecell{59}{65}
& \scorecell{59}{85}
& \deltacell{+}{6}{92}
& \deltacell{+}{4}{60}
& \scorecell{35}{20}
& \scorecell{40}{28}
& \deltacell{-}{0}{31}
& \deltacell{+}{0}{34} \\

& PASTA-GoE
& \scorecell{52}{82}
& \scorecell{55}{14}
& \deltacell{+}{0}{09}
& \deltacell{-}{0}{11}
& \scorecell{35}{63}
& \scorecell{39}{17}
& \deltacell{+}{0}{12}
& \deltacell{-}{0}{77} \\

& PASTA-$\cap$
& \scorecell{57}{65}
& \scorecell{59}{11}
& \deltacell{+}{4}{92}
& \deltacell{+}{3}{86}
& \scorecell{35}{99}
& \scorecell{39}{80}
& \deltacell{+}{0}{48}
& \deltacell{-}{0}{14} \\

& PASTA+SIAB
& \bscorecell{62}{76}
& \bscorecell{63}{09}
& \bdeltacell{+}{10}{03}
& \bdeltacell{+}{7}{84}
& \scorecell{37}{97}
& \scorecell{41}{61}
& \deltacell{+}{2}{46}
& \deltacell{+}{1}{67} \\

\cmidrule(lr){2-10}

& Random
& \scorecellpm{52}{39}{2.30}
& \scorecellpm{55}{27}{0.70}
& \deltacell{-}{0}{34}
& \deltacell{+}{0}{02}
& \scorecellpm{34}{64}{0.80}
& \scorecellpm{39}{75}{1.00}
& \deltacell{-}{0}{87}
& \deltacell{-}{0}{19} \\

\midrule
\addlinespace[2pt]

\multirow{11}{*}{%
  \modelcell{LLaMA3}{\citep{grattafiori2024llama3}}%
}

& ZS
& \scorecell{59}{27}
& \scorecell{60}{28}
& \dashcell
& \dashcell
& \scorecell{30}{39}
& \scorecell{33}{24}
& \dashcell
& \dashcell \\

& ZS+SIAB
& \scorecell{61}{80}
& \scorecell{63}{23}
& \deltacell{+}{2}{53}
& \deltacell{+}{2}{95}
& \scorecell{33}{64}
& \bscorecell{38}{02}
& \deltacell{+}{3}{25}
& \bdeltacell{+}{4}{78} \\

& Mid
& \scorecell{62}{24}
& \scorecell{62}{86}
& \deltacell{+}{2}{96}
& \deltacell{+}{2}{58}
& \scorecell{29}{16}
& \scorecell{34}{28}
& \deltacell{-}{1}{23}
& \deltacell{+}{1}{04} \\

& Mid+SIAB
& \scorecell{64}{21}
& \scorecell{64}{52}
& \deltacell{+}{4}{94}
& \deltacell{+}{4}{25}
& \scorecell{29}{81}
& \scorecell{36}{44}
& \deltacell{-}{0}{58}
& \deltacell{+}{3}{20} \\

& High
& \scorecell{64}{97}
& \scorecell{64}{81}
& \deltacell{+}{5}{70}
& \deltacell{+}{4}{53}
& \scorecell{26}{57}
& \scorecell{32}{53}
& \deltacell{-}{3}{82}
& \deltacell{-}{0}{72} \\

& High+SIAB
& \bscorecell{66}{22}
& \bscorecell{66}{13}
& \bdeltacell{+}{6}{95}
& \bdeltacell{+}{5}{85}
& \scorecell{27}{55}
& \scorecell{35}{85}
& \deltacell{-}{2}{84}
& \deltacell{+}{2}{61} \\

\cmidrule(lr){2-10}

& PASTA-ECC
& \scorecell{65}{28}
& \scorecell{63}{35}
& \deltacell{+}{6}{01}
& \deltacell{+}{3}{07}
& \scorecell{23}{20}
& \scorecell{26}{31}
& \deltacell{-}{7}{19}
& \deltacell{-}{6}{93} \\

& PASTA-GoE
& \scorecell{62}{09}
& \scorecell{61}{74}
& \deltacell{+}{2}{82}
& \deltacell{+}{1}{46}
& \scorecell{32}{14}
& \scorecell{31}{51}
& \deltacell{+}{1}{75}
& \deltacell{-}{1}{73} \\

& PASTA-$\cap$
& \scorecell{63}{28}
& \scorecell{60}{82}
& \deltacell{+}{4}{01}
& \deltacell{+}{0}{54}
& \scorecell{30}{86}
& \scorecell{33}{18}
& \deltacell{+}{0}{47}
& \deltacell{-}{0}{06} \\

& PASTA+SIAB
& \scorecell{65}{48}
& \scorecell{63}{55}
& \deltacell{+}{6}{21}
& \deltacell{+}{3}{27}
& \bscorecell{36}{17}
& \scorecell{37}{73}
& \bdeltacell{+}{5}{78}
& \deltacell{+}{4}{49} \\

\cmidrule(lr){2-10}

& Random
& \scorecellpm{52}{84}{5.00}
& \scorecellpm{56}{24}{4.10}
& \deltacell{-}{6}{43}
& \deltacell{-}{4}{04}
& \scorecellpm{28}{52}{2.80}
& \scorecellpm{30}{59}{3.20}
& \deltacell{-}{1}{87}
& \deltacell{-}{2}{65} \\

\addlinespace[3pt]
\bottomrule
\end{tabular}
\end{table*}
\subsection{Baselines and Prompting Settings}
We compare the proposed SIAB against three baselines. The first is prompting: Zero-Shot (ZS) and fixed few-shot prompting
(Templates are introduced in Appendix~\ref{app:prompts}).

The second is PASTA-style \citep{zhang2024pasta} performance-profiled head selection, for which we 
report three variants: \textsc{PASTA-ECC} and \textsc{PASTA-GoE} are profiled 
on labeled ECC and GoEmotions dev-data, respectively, probing in-domain 
effectiveness and cross-dataset transfer, \textsc{PASTA-$\cap$} keeps only 
the heads shared by both profiles, testing whether a shared profile improves 
generality. We also report \textsc{PASTA+SIAB} to test complementarity with in-domain PASTA. PASTA’s head count and variants are also tuned to their optimal values.

The third is, Layer-Matched Random, which tests whether the effect is only
layer-specific or due to within-layer redundancy. In either case, random
same-layer heads should work as well~\citep{voita-etal-2019-analyzing,michel2019sixteen}. We sample 20 random subsets of non-selected heads from the same layers, matching our head-set size, and tune $\beta$ on the same dev set.

\subsection{Results}
\label{sec:main-results}

Table~\ref{tab:merged-results} reports results on two datasets.
The reported changes in macro-F1 and full accuracy ($\Delta$) are computed relative to the ZS baseline.

\paragraph{\textsc{SIAB} generalized across models, datasets, and prompts.}
Across all 18 model--prompt--dataset settings, the proposed \textsc{SIAB} improved over
the corresponding prompting baseline using the same head set, with gains up
to $+10.27$ macro-F1 on the ECC benchmark and $+5.47$ on the GoEmotions benchmark. The effect is not
limited to ZS, but under 10-shot ECC, \textsc{SIAB} still added
$+1.21$ to $+4.25$ macro-F1 across models. On GoEmotions, it also reduced
harmful few-shot effects. It turned the Gemma3 model's Mid / High Top-1 drops into gains
and weakened the LLaMA3 model's Macro-F1 drops, while flipping its High-setting Top-1
change from negative to positive.
\paragraph{PASTA was strong in-domain but transferred poorly.}
In-domain PASTA profiles were competitive with, and often stronger than,
ZS+\textsc{SIAB}. However, this strength was task-specific. Applying
\textsc{PASTA-ECC} heads to GoEmotions dropped LLaMA3 by $-7.19$ macro-F1,
and transfer in the reverse direction was similarly unstable.
\textsc{PASTA-$\cap$} also lost in-domain performance without recovering
cross-task gains. In contrast, \textsc{SIAB} uses the same causally selected
head set across both datasets and gave positive $\Delta$ in all 18
settings. Thus, PASTA reached a stronger in-domain peak when its profiling
task matches the test task, whereas \textsc{SIAB} was weaker but more stable
across tasks, motivating their composition.

\paragraph{The two signals composed additively in five of six (model, dataset) configurations.}
Across five configurations, \textsc{PASTA+SIAB} recovered $79$--$116\%$ of the naive sum of the two individual ZS gains, and yielded the best macro-F1 on 4 of them, indicating the two signals operate largely orthogonally. The single exception, LLaMA3 on ECC, recovered only $54\%$. We attribute this to layer-band disjointness between the PASTA-profiled and SIAB head sets and quantify the relationship in \S\ref{sec:layer_band}.

\begin{figure*}
  \centering
  \includegraphics[width=\linewidth]{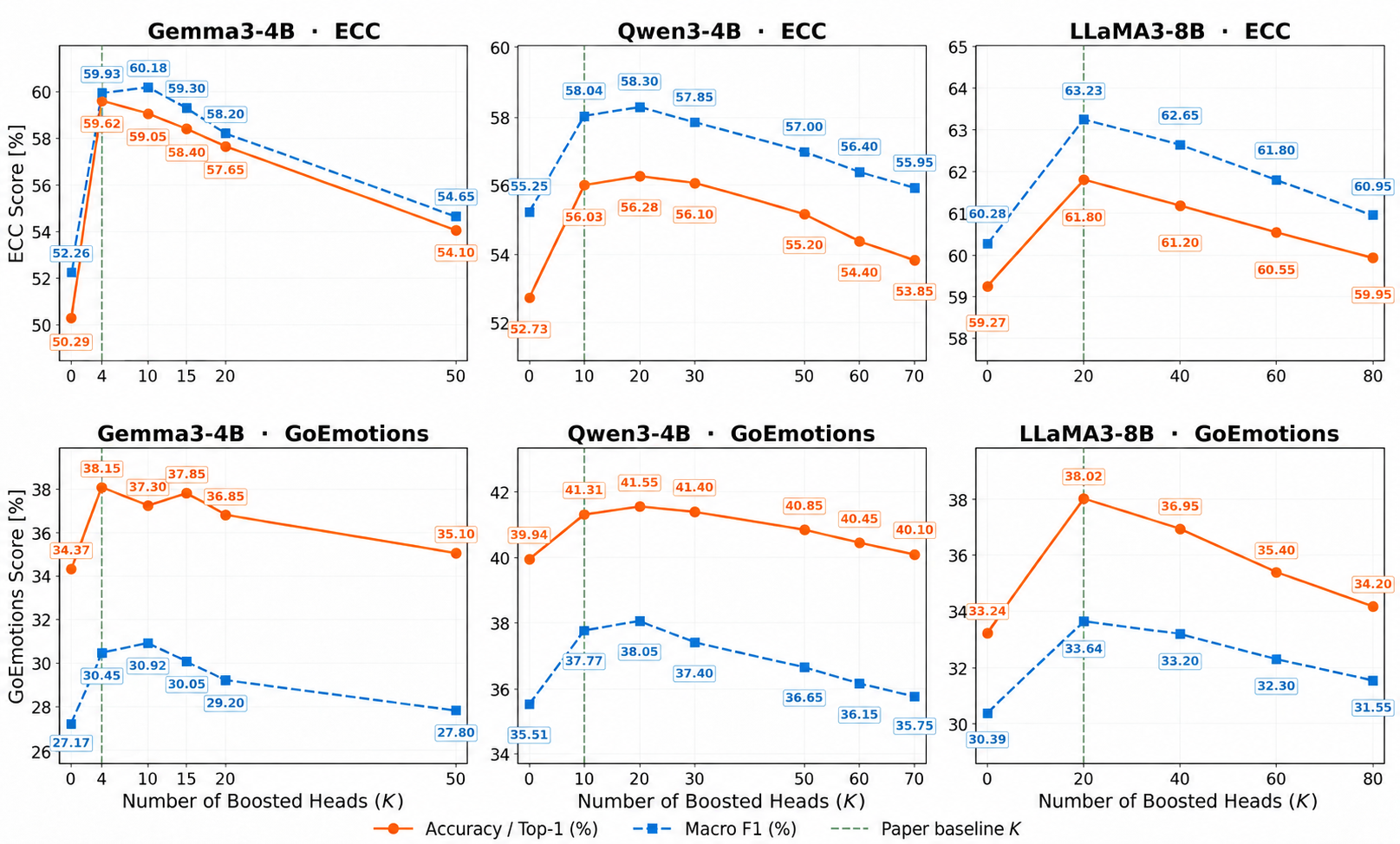}
\caption{Performance vs. boosted head count $K$ on the ECC~\citep{he2025ecc} and GoEmotions~\citep{demszky2020goemotions} benchmarks. 
Blue denotes macro-F1, and orange denotes accuracy for ECC and Top-1 accuracy for GoEmotions. 
Dashed lines mark the $K^{*}$ used in the main experiments, showing small-$K^{*}$ saturation across datasets for each model.}
  \label{fig:sparsity}
\end{figure*}

\paragraph{Effect of head selection: Evidence from Random.}
Across all models, datasets, and metrics, ZS+\textsc{SIAB} outperformed the mean score of 20 layer-matched random head subsets. Since Random uses the same
layers, head-set size, and tuning protocol, this argues against a simple
accumulation of layer-level effects. It also shows that same-layer heads are not
interchangeable for this intervention, but the gain depends on the specific heads
selected by SIAB.

\section{Mechanistic Analysis}
To analyze when and why the intervention succeeded, we partition test 
samples into four groups based on the joint behavior of ZS 
and the proposed SIAB predictions: 
\textbf{POS}: ZS incorrect, SIAB correct,
\textbf{NEG}: ZS correct, SIAB incorrect,
\textbf{EASY}: both ZS and SIAB correct, and
\textbf{HARD}: both ZS and SIAB incorrect.

\subsection{Sparsity of the Affective Head Set}
\label{sec:sparsity}
To test whether our selected head set is sufficiently sparse, we vary $K$ and select the top-$K$ heads according to the ranking determined during refinement in \S\ref{sec:refinement}.
Figure~\ref{fig:sparsity} shows the same trend across all models and datasets, that
F1 rose quickly with a few heads, saturated near a small $K^{*}$, and dropped
as more heads were added. This suggests that affective signal is concentrated
in a small head set, while lower-ranked heads mostly add noise. Since the
same $K^{*}$ appears on both the ECC and GoEmotions benchmarks for each model, the 
sparsity seems more model-dependent than dataset-specific.

\subsection{Selected Heads Attend More to Emotion Words}
\label{sec:emotion_attention}

We test whether selected heads disproportionately attend to emotion words.
\paragraph{Setup.}
For each dialogue turn $u$ that contains at least one word from the
NRC Emotion Lexicon~\citep{mohammad2013crowdsourcing}, let
$\mathcal{C}_u$ denote the content-word tokens in the turn, and let
$\mathcal{E}_u \subseteq \mathcal{C}_u$ denote the subset of tokens matched by
the NRC. For a head $h$, let $\mathcal{S}_{h,u}$ be the top-$30\%$
tokens in $\mathcal{C}_u$ ranked by the attention that head $h$ assigns to
them at the prediction position. We then compute a concentration ratio defined as:
\begin{equation}
r_{h,u}
=
\frac{|\mathcal{S}_{h,u}\cap\mathcal{E}_u|/|\mathcal{S}_{h,u}|}
     {|\mathcal{E}_u|/|\mathcal{C}_u|}.
\label{eq:concentration_ratio}
\end{equation}
The numerator is the emotion-word fraction among the head's top-attended
tokens, while the denominator is the emotion-word fraction in the whole turn.
Thus, $r_{h,u}>1$ indicates that head $h$ attends to emotion words more than
expected from their base rate in that turn. We compare selected heads with
random non-selected heads sampled from early, middle, and late layer bands
($40$ heads per band).
\paragraph{Internal signal was head-specific, but not layer-driven.}
The selected heads had the highest emotion-word ratios for all three models as shown in
Table~\ref{tab:attention}. Among random heads, middle and late layer bands
generally showed higher ratios than early layers, consistent with prior work
showing stronger affective or sentiment signals in mid-to-late
layers~\citep{tak2025mechanistic,tigges2024sentiment}. Moreover, the selected heads span
early, middle, and late layer regions, yet outperformed random heads within each
region. This suggests that the higher emotion-word ratio is a property of the selected
heads themselves, not merely a consequence of their layer positions. Appendix~\ref{sec:hard_discussion} provides qualitative examples.
\begin{table}[t]
\centering
\caption{Emotion-word ratio for selected and random heads.
Larger values indicate a higher emotion-word share turn average.}
\label{tab:attention}
\small
\setlength{\tabcolsep}{4pt}
\begin{tabularx}{\columnwidth}{@{}l*{3}{C}@{}}
\toprule
\multirow{2}{*}{\textbf{Head set}}
& \textbf{Gemma3} & \textbf{Qwen3} & \textbf{LLaMA3} \\
& {\footnotesize ($K{=}4$)}
& {\footnotesize ($K{=}10$)}
& {\footnotesize ($K{=}20$)} \\
\midrule
Selected            & \textbf{1.575} & \textbf{1.370} & \textbf{1.384} \\
\midrule
Random Late         & 1.373 & 1.064 & 1.195 \\
Random Middle       & 1.346 & 1.157 & 1.228 \\
Random Early        & 1.224 & 1.083 & 1.092 \\
\bottomrule
\end{tabularx}
\end{table}

\begin{table}[t]
\centering
\caption{Mean $\Delta\text{label\_gap}$ per on GoEmotion samples.}
\label{sec:emotion_attention}
\small
\begin{tabular}{lccc}
\toprule
\textbf{Group} & \textbf{Gemma3} & \textbf{Qwen3} & \textbf{LLaMA3} \\
\midrule
POS  & $+9.77$ & $+2.85$ & $+0.79$ \\
NEG  & $-7.57$ & $-2.12$ & $-0.83$ \\
HARD & $+2.53$ & $+1.14$ & $+0.45$ \\
\bottomrule
\end{tabular}
\end{table}

\begin{table}[t]
\centering

\caption{Zero-Shot (ZS) margin and label-gap statistics by sample group, and rescue rates by label-gap range.In (a), share is calculated over all test samples, and in (b), it is calculated over the POS+HARD subset.}
\label{tab:logit_geometry}

\small
\setlength{\tabcolsep}{4pt}

\begin{subtable}{\columnwidth}
\caption{ZS margin =
$\ell_{\mathrm{top1}}-\ell_{\mathrm{top2}}$ and ZS label\_gap =
$\ell_{\mathrm{gold}}-\ell_{\mathrm{top1}}$ by group.}
\label{tab:logit_geometry_a}

\begin{tabular*}{\linewidth}{@{\extracolsep{\fill}}lrrr@{}}
\toprule
\textbf{Group} & \shortstack[c]{\textbf{Share [\%]}} & \textbf{ZS margin} & \textbf{ZS label\_gap} \\
\midrule
POS  & $12.9$ & $9.4$  & $-9.8$  \\
NEG  & $3.1$  & $8.7$  & $0.0$  \\
EASY & $42.7$ & $20.5$ & $0.0$  \\
HARD & $41.3$ & $16.8$ & $-20.2$ \\
\bottomrule
\end{tabular*}
\end{subtable}

\vspace{0.5em}

\newcommand{\range}[2]{%
  $[\makebox[2.5em][r]{$#1$},\makebox[1.6em][r]{$#2$}]$%
}

\begin{subtable}{\columnwidth}
\caption{Rescue rate of POS+HARD samples stratified by ZS label\_gap.}
\label{tab:logit_geometry_b}

\begin{tabular*}{\linewidth}{@{\extracolsep{\fill}}lrrr@{}}
\toprule
\textbf{Range} & \shortstack[c]{\textbf{Share [\%]}} & \multicolumn{2}{c}{\textbf{Rescue rate}} \\
\midrule
\range{-5}{0} (Shallow)    & $10.2$ & \multicolumn{2}{c}{$\mathbf{66.9\%}$} \\
\range{-10}{-5} (Mid)      & $12.3$ & \multicolumn{2}{c}{$47.9\%$} \\
\range{-100}{-10} (Deep)   & $77.5$ & \multicolumn{2}{c}{$14.4\%$} \\
\bottomrule
\end{tabular*}
\end{subtable}

\end{table}

\begin{table}[t]
\centering
\caption{Selected-head attention mass on emotion tokens. POS shows stronger label-directional bias.}
\label{tab:nrc_alignment}
\small
\setlength{\tabcolsep}{8pt}
\begin{tabular}{@{}lccc@{}}
\toprule
\textbf{Group} & \textbf{$r_{\text{label}}$} & \textbf{$r_{\text{wrong}}$} & \textbf{$\Delta r$} \\
\midrule
POS  & $0.767 \pm 0.284$ & $0.140 \pm 0.233$ & $+0.627$ \\
NEG  & $0.661 \pm 0.335$ & \multicolumn{1}{c}{\textemdash} & \multicolumn{1}{c}{\textemdash} \\
EASY & $0.726 \pm 0.325$ & \multicolumn{1}{c}{\textemdash} & \multicolumn{1}{c}{\textemdash} \\
HARD & $0.656 \pm 0.345$ & $0.195 \pm 0.285$ & $+0.461$ \\
\bottomrule
\end{tabular}
\end{table}

\subsection{What Makes a Sample Rescuable?}
\label{sec:predicts_rescue}

The proposed SIAB changes the prediction only when its induced logit movement is both
large enough and pointed in the useful direction. We first describe the
observed movement across POS, NEG, and HARD samples, and then analyze two
factors behind it: Baseline logit gap and Alignment of selected heads
with the gold label.

\paragraph{SIAB induced different logit movements across groups.}
We measure the intervention effect by
$\Delta\text{label\_gap} = \text{label\_gap}_{\text{SIAB}} -
\text{label\_gap}_{\text{ZS}}$. For an example with gold label $y$, candidate label set $\mathcal{Y}$,
and label-logit vector $\ell$, we define
$\text{label\_gap}=\ell_y-\max_{c\in\mathcal{Y}}\ell_c$, where $c$
indexes candidate emotion labels. The gap is $0$ when the gold label is
already Top-1 and negative otherwise, and positive
$\Delta\text{label\_gap}$ values mean that SIAB moves the gold label
closer to Top-1. As reported in Table~\ref{sec:emotion_attention}, POS shifted strongly upward and NEG strongly downward. Meanwhile, HARD also shifted upward,
but by a much smaller margin.
\paragraph{Large baseline logit gaps limited rescue.}
The weaker HARD movement is partly explained by its baseline logit geometry.
Table~\ref{tab:logit_geometry}\,(a) shows that POS and NEG had small Top-1
vs.\ Top-2 margins on the Gemma3 model ($9.4$ and $8.7$), while HARD had a much larger
margin ($16.8$) and a deeper gold-label gap ($-20.2$). Stratifying POS+HARD
samples by ZS label\_gap further confirms this pattern as shown in 
Table~\ref{tab:logit_geometry}\,(b): Rescue was frequent in the shallow range
but dropped sharply once the gold label was more than $10$ logit points behind
Top-1. HARD mostly lay in this deep-gap regime, where SIAB could move the gold
label but rarely enough to change the prediction.

\paragraph{Direction was set by head alignment.}
Selected
heads differed across groups in how strongly they attended to gold-label versus
wrong-label emotion tokens before intervention. Using the NRC EmoLex categories, we compute
the fraction of selected-head attention mass on emotion tokens matching the gold
label ($r_{\text{label}}$) versus the wrong Top-1 ($r_{\text{wrong}}$). As shown in Table 5, $r_{\text{label}} > 0.65$. POS
exhibited stronger directional bias than HARD ($0.767$ vs.\ $0.656$) and a lower
wrong-aligned fraction ($0.140$ vs.\ $0.195$). On HARD, the intervention thus
amplifies a noisier signal, which can partially explain
HARD's resistance to rescue.

\paragraph{Can more HARD samples be rescued?}
We probe this in two ways: by scaling $\beta$ by a factor of 2 and by suppressing
sink-token attention in selected heads
(Table~\ref{tab:hard_sink_suppression_full}). First, setting $\beta$ to
$2\beta$ rescued only $4.5\%$ of HARD samples on Gemma3, and the effect
saturated beyond $\times 2$. LLaMA3 and Qwen3 showed essentially no
gain ($0.1\%$ and $0.4\%$, respectively). Thus, increasing $\beta$ alone
is insufficient. Second, suppressing attention to sink tokens in selected heads generally
performed better than increasing $\beta$, but the gains remained modest
despite the large share of attention attracted by sink tokens.
The largest gain on HARD samples was $5.5\%$ for LLaMA3, while Qwen3
showed only a small gain.

\begin{table}[t]
\centering
\caption{Rescue rate on HARD samples under two intervention strategies.}
\label{tab:hard_sink_suppression_full}
\small
\setlength{\tabcolsep}{4pt}
\begin{tabular}{lccc}
\toprule
\multirow{2}{*}{\textbf{Model}}
& \textbf{Sink}
& \multirow{2}{*}{\textbf{$\boldsymbol{\beta}\times 2$}}
& \textbf{Sink} \\
& \textbf{mass}
&
& \textbf{suppression} \\
\midrule
Gemma3 & $\sim\!25\%$ & $4.5\%$ & $4.2\%$ \\
Qwen3  & $\sim\!40\%$ & $0.4\%$ & $1.4\%$ \\
LLaMA3 & $\sim\!60\%$ & $0.1\%$ & $5.5\%$ \\
\bottomrule
\end{tabular}
\end{table}

\subsection{Layer-band Proximity in SIAB–PASTA Additivity}
\label{sec:layer_band}

The Results section identified the LLaMA3 model on the ECC benchmark as the single non-additive 
outlier. Here we test whether this outlier was caused by \emph{layer 
disjointness} between the two head sets.

\paragraph{Quantifying additivity.}
For each PASTA head set, we compare the actual Combo gain to the sum of the 
two individual gains as:
\begin{equation}
  \text{Additivity}
  =
  \frac{\Delta_{\text{Combo}}}
       {\Delta_{\text{PASTA}} + \Delta_{\text{SIAB}}},
\label{eq:additivity}
\end{equation}
where $\Delta_X = X - \text{ZS}$. A value of $100\%$ means the two 
interventions are exactly additive; $<100\%$ indicates redundancy or 
interference between the two head sets and $>100\%$ indicates super-additivity.

\paragraph{Layer-disjointness hypothesis.}
We hypothesize that PASTA and the proposed SIAB compose additively only when the two head 
sets occupy overlapping or adjacent Transformer layers, and breaks down when 
the two sets sit too many layers apart for their signals to combine.
LLaMA3 on ECC was the natural witness. Its scan-selected PASTA Top-10 
lay entirely on $L{\leq}11$, and Combo provided essentially nothing on top 
of PASTA alone. To probe the causal direction, we (i) re-select LLaMA3's 
PASTA heads to mid-late layers and check whether additivity recovers, and 
(ii) artificially reproduce the disjoint pattern in two models where Combo 
otherwise works by hand-selecting 
their PASTA-positive heads at near-SIAB vs.\ early layer bands.

\paragraph{Manipulating PASTA's layer band recovered and broke additivity.}
Table~\ref{tab:layer_band} reports additivity per manipulation, and the 
pattern was consistent across all three models. Additivity was high when 
PASTA and SIAB occupied nearby layers and collapsed when they were forced 
apart. Re-selecting LLaMA3's natively early-stacked PASTA heads 
 $(\bar{d}{=}8.6)$ to mid-late layers $(\bar{d}{=}0.3)$ lifted F1 additivity 
above $100\%$ ($108.0\%$). Conversely, forcing the otherwise-additive 
Gemma3 and Qwen3 models' PASTA heads into early layers dropped additivity from 
$95.8\%$ and $118.5\%$ (Near) to $79.4\%$ and $70.0\%$ (Far). The recovery 
and breakdown both track the layer gap rather than the model, confirming 
that the layer band is the limiting factor.

\begin{table}[t]
\centering
\caption{Layer-band manipulation on the ECC~\citep{he2025ecc} benchmark. \emph{Near} / \emph{Far} denote 
PASTA \citep{zhang2024pasta} heads at small / large mean layer distance $\bar{d}$ to the proposed SIAB's heads, respectively. 
Boldface values mark the high-additivity configuration per model. 
$^{\dagger}$Configuration used in the main results (Table~\ref{tab:merged-results}).}
\label{tab:layer_band}
\small
\setlength{\tabcolsep}{3pt}

\begin{tabular}{@{}llr@{\hspace{3pt}}c@{\hspace{1pt}}r@{}}
\toprule
\textbf{Model} 
& \textbf{PASTA dist.\ to SIAB} 
& \multicolumn{3}{c}{\textbf{Additivity (Acc / F1)}} \\
\midrule

Gemma3 
& Near $(\bar{d}{=}1.1)$
& $\mathbf{95.8\%}$ &  / & $\mathbf{96.6\%}$ \\

Gemma3 
& Far $(\bar{d}{=}10.5)$
& $79.4\%$ &  / & $81.5\%$ \\

\midrule

Qwen3 
& Near $(\bar{d}{=}0.5)$
& $\mathbf{118.5\%}$ &  / & $\mathbf{112.7\%}$ \\

Qwen3 
& Far $(\bar{d}{=}11.1)$
& $70.0\%$ &  / & $70.4\%$ \\

\midrule

LLaMA3 
& Near $(\bar{d}{=}0.3)$
& $\mathbf{89.0\%}$ &  / & $\mathbf{108.0\%}$ \\

LLaMA3 
& Far $(\bar{d}{=}8.6)$$^{\dagger}$
& $72.7\%$ &  / & $54.3\%$ \\

\bottomrule
\end{tabular}
\end{table}

\section{Conclusion}
Affective state in dialogue depends on contextual cues that are already present but easily lost at prediction time. We showed that the proposed SIAB could route this evidence through internal model signals, without relying on externally marked spans or task-specific training. A fixed causally localized head set transferred across two contrasting attentional demands, whereas PASTA-style external profiles were competitive in-domain but transferred poorly. Their additive gains further suggested that internal and external signals capture complementary aspects of affective evidence. Future work should move from a fixed global intervention to sample-adaptive control, where intervention strength and target tokens vary with each example's uncertainty and attention pattern, and test whether the same interface extends to other latent conversational states such as uncertainty, trust, or intent.


\section*{Limitations}

\paragraph{Model scale.}
We evaluated three open instruction-tuned LLMs spanning $4$--$8$B 
parameters from distinct families (Gemma3, Qwen3, and LLaMA3). Whether the 
internal signal continues to generalize at much larger scales 
(e.g., $70$B$+$) is an open question. Exhaustive activation patching at such 
scales is computationally prohibitive in our setting.

\paragraph{Language coverage.}
Both benchmarks (ECC and GoEmotions) were in English. Whether the recovered 
head set transfers across languages, or whether language-specific 
recognition circuits would emerge under multilingual evaluation, is 
untested.

\paragraph{Dataset breadth.}
The two benchmarks were selected to span complementary attentional 
demands (long-context dialogue integration vs.\ wide-label distribution 
shift over short inputs), and we reuse the same head set across them 
without re-derivation. Generalization to non-emotion classification 
tasks is left to future work.

\paragraph{Boundary on HARD samples.}
A subset of test samples (HARD) remained unrescued even under stronger 
interventions. As analyzed in \S\ref{sec:predicts_rescue}, Both 
scaling $\beta$ and suppressing attention sink were insufficient. Whether these samples can be rescued by a more selective, 
sample-adaptive intervention is left to future work.
\bibliography{custom}

\appendix

\begin{table*}[t]
\centering
\caption{Hyperparameter inventory. The first six are fixed per model
across all (dataset, shot) configurations, while the last one is recalibrated as noted.}
\label{tab:hparam_inventory}

\small
\setlength{\tabcolsep}{4pt}
\renewcommand{\arraystretch}{1.1}
\begin{tabular}{llll}
\toprule
Hyperparameter & Role & Value & How set \\
\midrule
$\tau_\text{head}$ & Direct-effect threshold & 0.05 & \citep{wang2023ioi} \\
$\lambda$ & Attention-vs.-value weight in $s_\text{head}$& 0.5 & Equal-weighting \\
$K$ & Heads in $\mathcal{H}$ & 4, 10, 20 & $K^*$ saturation, Fig.~\ref{fig:sparsity} \\
$|\mathcal{Q}|$ & Boost-applied query positions & 4 & ECC dev \\
$r$ & Top fraction of attended content tokens & 0.3 & A priori \\
$\beta$ & Pre-softmax bias magnitude & Tab~\ref{tab:selected_heads}& Per model \\
$(k, t)$ & GoEmotions multi-label top-k + threshold & Tab~\ref{tab:Tt_calibration}& Per (model, shot), baseline only \\
\bottomrule
\end{tabular}
\end{table*}

\section{Pair Construction for Causal Patching}
\label{app:pair_construction}

Following the minimal-pair design of~\citet{wang2023ioi}, each pair
$(X^{\mathrm{clean}}, X^{\mathrm{corr}})$ shares syntactic skeleton
and token length and differs only in affective content.

\paragraph{Coverage.}
The set spans the six basic emotions of~\citet{ekman1992argument}:
Anger, Disgust, Fear, Happiness, Sadness, and Surprise, 100 pairs each
(600 total), covering both valences and arousal levels to avoid a
polarity confound. Example (Anger): clean \texttt{``During class, you
blamed me for the result, and I feel very''} (Target \texttt{angry})
vs.\ corrupted \texttt{``\dots you praised me\dots''} (Target
\texttt{happy}). Within each category, pairs further vary the
syntactic locus of the cue ---verb, predicate-adjective,
prenominal-adjective, or purpose-clause replacement--- and distribute
it across multiple content words, so the discovered head set is not
tied to a single syntactic position.

\paragraph{Verification.}
We constructed these datasets from approximately 25 templates authored by human annotators. Each template contains several replaceable slots, including names, times, locations, objects, and an emotion-bearing word or phrase. The emotion-bearing slots are shown in bold in the full template list. Example templates include:

\begin{itemize}
\item ``[time PP], [Name] \textbf{[verb-past]} me in front of [audience NP], which makes me feel''
    \item ``I received a \textbf{[adj]} [noun] from [person NP], and I feel''
\item ``The \textbf{[adj]} [noun] appeared [time/place PP], which makes me feel''
    \item ``The results of my [noun] came back \textbf{[adj]} [time PP], and I feel''
\item ``[Name] stayed up all night to \textbf{[VP]}, and I feel''
\end{itemize}

Mistral-Small-3.2-24B-Instruct-2506~\citep{mistral_small_3_2_2025}, which is distinct from all downstream models evaluated in this work, instantiated these templates with concrete names, times, locations, objects, and emotion-bearing expressions.

\paragraph{Generalization beyond localization categories.}
Only 4 of 10 labels in the ECC benchmark and 6 of 27 labels in the GoEmotions benchmark overlap with
the pair-set categories. The rest are unseen during localization.
Strong performance on these unseen labels as shown in \S\ref{sec:main-results}, 
indicates that the head set generalizes beyond the specific cues
used to identify it.

\section{Implementation Details for Refinement and Intervention}
\label{app:refinement}

Table~\ref{tab:hparam_inventory} summarizes the complete set of hyperparameters used by the proposed SIAB, listing the role, value, and how each is set. The per-model final head set $\mathcal{H}$ and the calibrated intervention strength $\beta$ are reported in Table~\ref{tab:selected_heads}.

\paragraph{Non-content token filter and heads detail.}
A token is excluded from $A_h, V_h$ averaging and from the
intervention pool $\mathcal{T}_r(h,q)$ if it falls into any of
(i) punctuation, (ii) English personal pronouns,
(iii) interrogatives (\textit{what, who, when, where, why, how,
which}), (iv) prepositions, or (v) model-specific special tokens
(BOS / EOS and chat-template markers). Negation and intensifiers
(\textit{not, very, no, \dots}) are retained as content tokens.

\begin{table}[t]
\centering
\caption{Final head set $\mathcal{H}$ per model, with intervention
strength $\beta$. Both $\mathcal{H}$ and $\beta$ are fixed
per model and reused across the two datasets. All indices are 0-based. HuggingFace
identifiers: \texttt{google/gemma-3-4b-it},
\texttt{Qwen/Qwen3-4B-Instruct-2507},
\texttt{meta-llama/Llama-3.1-8B-Instruct}.}
\label{tab:selected_heads}
\small
\setlength{\tabcolsep}{3pt}
\renewcommand{\arraystretch}{1.15}
\begin{tabularx}{\columnwidth}{@{}>{\raggedright\arraybackslash}p{0.32\columnwidth}>{\raggedright\arraybackslash}X@{}}
\toprule
\textbf{Model} & \textbf{Selected (layer, head) indices} \\
\midrule
Gemma3 \mbox{($K{=}4$, $\beta{=}6$)}
& (24,4), (17,2), (22,6), (17,0) \\
\midrule
Qwen3 \mbox{($K{=}10$, $\beta{=}9$)}
& (22,5), (18,30), (23,10), (20,9), (29,3),
  (26,27), (29,0), (7,3), (25,9), (24,15) \\
\midrule
LLaMA3 \mbox{($K{=}20$, $\beta{=}10$)}
& (18,0), (16,10), (19,9), (16,26), (21,26),
  (6,20), (22,1), (13,10), (15,26), (21,24),
  (15,14), (19,11), (14,24), (15,20), (20,9),
  (13,12), (17,28), (24,21), (28,25), (17,20) \\
\bottomrule
\end{tabularx}
\end{table}

\section{Hyperparameters and Calibration}
\label{app:hparams}
This section describes how the calibration sets are constructed and how each hyperparameter is tuned or frozen. The multi-label decoding parameters $(k,t)$ used for each (model, shot) configuration for the GoEmotions benchmark are listed in Table~\ref{tab:Tt_calibration}.

\paragraph{Calibration sets.}
 For the ECC benchmark, 200 dialogues sampled from the official dev split form the calibration set, while the evaluation uses the full ECC test split. For the GoEmotions benchmark, 200 dev samples form the calibration set, while the evaluation uses the full GoEmotions test split, excluding Neutral-only samples. The per-model $\beta$ is calibrated once on the ECC dev set and reused unchanged on GoEmotions. No hyperparameter is tuned on test data.

\paragraph{Hyperparameter inventory.}
The structural hyperparameters ($\tau_\text{head}, \lambda, K, |\mathcal{Q}|$ and $r$) are calibrated once on the ECC dev set and frozen across all (model, dataset, shot) configurations. The bias magnitude $\beta$ is likewise calibrated once per model on the ECC dev set and held fixed across both datasets. $|\mathcal{Q}|{=}4$ tokens around the answer position, $r{=}0.3$ of content-word tokens, $K{=}4, 10, 20$ for the Gemma3, Qwen3, and LLaMA3 models, respectively, set at each model's saturation point on the ECC dev sweep (Figure~\ref{fig:sparsity}), $\beta{=}6, 9, 10$ for Gemma3, Qwen3, and LLaMA3, respectively. The GoEmotions multi-label decoding uses a Top-$T$ cap and inclusion threshold $t$, tuned on the GoEmotions dev set under the baseline and frozen for \textsc{+SIAB}, per-(model, shot) values are shown in Table~\ref{tab:Tt_calibration}.

\begin{figure*}[t]
\centering
\includegraphics[width=\textwidth]{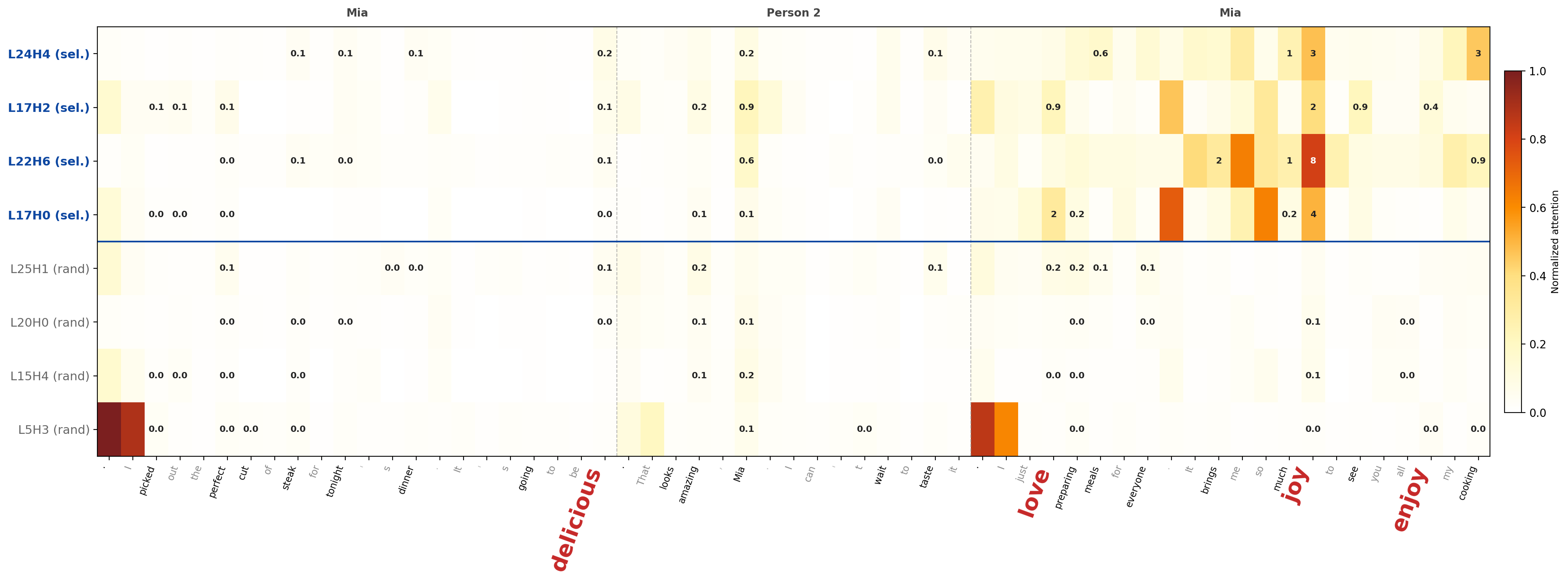}
\caption{Per-head attention on an ECC~\citep{he2025ecc} example rescued by the proposed SIAB
(GT \emph{happy}, ZS\,$\to$\,\emph{proud}~\ding{55},
ZS+SIAB\,$\to$\,\emph{happy}~\ding{51}). Each row: Attention from
the final-4 query band on a single head, globally normalized, cell
value $=$ attention $\times 100$, restricted to the head's
top-30\% content tokens (the boost set $\mathcal{S}_{h,q}$).
Bold red tokens are NRC EmoLex~\citep{mohammad2013crowdsourcing} emotion words. Top: Four
Gemma3~\citep{gemma2025technical} selected heads (L24H4, L17H2, L22H6, L17H0). Bottom:
Depth-stratified random non-selected heads (L25H1, L20H0, L15H4,
L5H3).}
\label{fig:hero}
\end{figure*}

\begin{table}[t]
\centering
\caption{Multi-label decoding parameters $(T,t)$ per
(model, shot) pair for the GoEmotions benchmark: $T$ = Temperature, $t$ = inclusion
threshold. Tuned on GoEmotions dev under baseline and frozen for
+SIAB, so reported $\Delta$F1 reflects intervention effect, not
$(T,t)$ re-calibration.}
\label{tab:Tt_calibration}
\small
\setlength{\tabcolsep}{6pt}
\renewcommand{\arraystretch}{1.1}
\begin{tabular}{r@{\hspace{12pt}}ccc}
\toprule
\multicolumn{1}{c}{\textbf{Shots}} & \textbf{Qwen3} & \textbf{LLaMA3} & \textbf{Gemma3} \\
\midrule
0  & ( 5, 0.10)& ( 3, 0.10)& ( 6, 0.05)\\
5  & ( 3, 0.20)& ( 3, 0.10)& ( 7, 0.05)\\
10 & ( 2, 0.20)& ( 3, 0.10)& (10, 0.05) \\
15 & ( 7, 0.08)& ( 3, 0.10)& ( 3, 0.03)\\
27 & ( 3, 0.15)& ( 2, 0.15)& ( 7, 0.05)\\
\bottomrule
\end{tabular}
\end{table}

\section{Results on Neutral-only GoEmotions Samples}
\label{app:neutral}
We exclude samples labeled solely as Neutral, since they lack
explicit emotion cues. Multi-label samples retain their non-Neutral
labels. On the excluded Neutral-only subset (1,606 test samples),
the proposed SIAB \emph{improved} Neutral recall by $+1.08$ to $+1.93$ percentage
points across the three models (Gemma3: 9.10\,$\to$\,10.54,
Qwen3: 10.05\,$\to$\,11.13, and LLaMA3: 8.93\,$\to$\,10.86),
indicating that content-token amplification does not inflate emotion
predictions on neutral inputs.

\section{Qualitative Attention Visualization}
\label{sec:hard_discussion}

Figure~\ref{fig:hero} visualizes per-head
attention on an ECC benchmark example,
contrasting the four selected heads (Top) from the Gemma3 model with four
depth-stratified random non-selected heads (Bottom).

\begin{table}[!t]
\centering
\caption{Pointwise paired-bootstrap 95\% percentile Confidence
Intervals (CI) on the GoEmotions~\citep{demszky2020goemotions} benchmark ($B=10{,}000$, values are $\times100$).
The proposed SIAB comparisons are prompt-matched, whereas PASTA-family rows
are compared with Zero-Shot (ZS).}
\label{tab:bootstrap-goe}

\scriptsize
\setlength{\tabcolsep}{2.5pt}
\renewcommand{\arraystretch}{0.90}

\begin{minipage}{0.82\linewidth}
\raggedright
\textbf{(a) Gemma3}~\citep{gemma2025technical}

\vspace{0.25em}

\begin{tabular}{@{}lcc@{}}
\toprule
\textbf{Comparison}
& \shortstack{\textbf{$\Delta$Top-1}\\{[95\% CI]}}
& \shortstack{\textbf{$\Delta$F1}\\{[95\% CI]}} \\
\midrule

ZS $\rightarrow$ ZS+\textsc{SIAB}
& \bootci{+3.78}{+2.93}{+4.73}
& \bootci{+3.28}{+2.16}{+4.42} \\
Mid $\rightarrow$ Mid+\textsc{SIAB}
& \bootci{+3.23}{+2.31}{+4.18}
& \bootci{+2.10}{+0.94}{+3.19} \\
High $\rightarrow$ High+\textsc{SIAB}
& \bootci{+3.95}{+3.03}{+4.89}
& \bootci{+2.77}{+1.65}{+3.86} \\
\textsc{PASTA-ECC} vs.\ ZS
& \bootci{-1.48}{-2.53}{-0.53}
& \bootci{-0.31}{-1.30}{+0.79} \\
\textsc{PASTA-GoE} vs.\ ZS
& \bootci{+0.00}{-0.93}{+1.11}
& \bootci{+4.64}{+3.46}{+5.67} \\
\textsc{PASTA}-$\cap$ vs.\ ZS
& \bootci{+0.28}{-0.72}{+1.35}
& \bootci{+2.02}{+0.96}{+3.22} \\
\textsc{PASTA+SIAB} vs.\ ZS
& \bootci{+3.64}{+2.70}{+4.77}
& \bootci{+6.58}{+5.57}{+7.76} \\

\bottomrule
\end{tabular}
\end{minipage}
\vspace{0.6em}

\begin{minipage}{0.82\linewidth}
\raggedright
\textbf{(b) Qwen3}~\citep{yang2025qwen3}

\vspace{0.25em}

\begin{tabular}{@{}lcc@{}}
\toprule
\textbf{Comparison}
& \shortstack{\textbf{$\Delta$Top-1}\\{[95\% CI]}}
& \shortstack{\textbf{$\Delta$F1}\\{[95\% CI]}} \\
\midrule

ZS $\rightarrow$ ZS+\textsc{SIAB}
& \bootci{+1.37}{+0.48}{+2.27}
& \bootci{+2.26}{+1.10}{+3.63} \\
Mid $\rightarrow$ Mid+\textsc{SIAB}
& \bootci{+2.31}{+1.48}{+3.26}
& \bootci{+2.75}{+1.41}{+4.06} \\
High $\rightarrow$ High+\textsc{SIAB}
& \bootci{+1.01}{+0.12}{+1.88}
& \bootci{+2.24}{+0.93}{+3.67} \\
\textsc{PASTA-ECC} vs.\ ZS
& \bootci{+0.34}{-0.61}{+1.32}
& \bootci{-0.31}{-1.59}{+0.96} \\
\textsc{PASTA-GoE} vs.\ ZS
& \bootci{-0.77}{-1.57}{+0.18}
& \bootci{+0.12}{-1.33}{+1.61} \\
\textsc{PASTA}-$\cap$ vs.\ ZS
& \bootci{-0.14}{-0.97}{+0.76}
& \bootci{+0.48}{-0.90}{+1.86} \\
\textsc{PASTA+SIAB} vs.\ ZS
& \bootci{+1.67}{+0.82}{+2.69}
& \bootci{+2.46}{+1.01}{+3.75} \\

\bottomrule
\end{tabular}
\end{minipage}
\vspace{0.6em}

\begin{minipage}{0.82\linewidth}
\raggedright
\textbf{(c) LLaMA3}~\citep{grattafiori2024llama3}

\vspace{0.25em}

\begin{tabular}{@{}lcc@{}}
\toprule
\textbf{Comparison}
& \shortstack{\textbf{$\Delta$Top-1}\\{[95\% CI]}}
& \shortstack{\textbf{$\Delta$F1}\\{[95\% CI]}} \\
\midrule

ZS $\rightarrow$ ZS+\textsc{SIAB}
& \bootci{+4.78}{+3.67}{+5.83}
& \bootci{+3.25}{+2.15}{+4.31} \\
Mid $\rightarrow$ Mid+\textsc{SIAB}
& \bootci{+2.16}{+1.07}{+3.23}
& \bootci{+0.65}{+0.03}{+1.29} \\
High $\rightarrow$ High+\textsc{SIAB}
& \bootci{+3.32}{+2.20}{+4.37}
& \bootci{+0.98}{+0.06}{+1.87} \\
\textsc{PASTA-ECC} vs.\ ZS
& \bootci{-6.93}{-8.13}{-5.80}
& \bootci{-7.19}{-8.31}{-6.08} \\
\textsc{PASTA-GoE} vs.\ ZS
& \bootci{-1.73}{-2.77}{-0.70}
& \bootci{+1.75}{+0.75}{+2.83} \\
\textsc{PASTA}-$\cap$ vs.\ ZS
& \bootci{-0.06}{-1.09}{+1.14}
& \bootci{+0.47}{-0.41}{+1.44} \\
\textsc{PASTA+SIAB} vs.\ ZS
& \bootci{+4.49}{+3.45}{+5.60}
& \bootci{+5.78}{+4.67}{+6.88} \\

\bottomrule
\end{tabular}
\end{minipage}
\end{table}

\begin{table}[!t]
\centering
\caption{Pointwise paired-bootstrap 95\% percentile Confidence
Intervals (CI) on the ECC~\citep{he2025ecc} benchmark ($B=10{,}000$, values are $\times100$).
The proposed SIAB comparisons are prompt-matched, whereas PASTA-family rows
are compared with Zero-Shot (ZS).}
\label{tab:bootstrap-ecc}
\vspace{\baselineskip}
\scriptsize
\setlength{\tabcolsep}{2.5pt}
\renewcommand{\arraystretch}{0.90}
\newcommand{\bootciecc}[3]{%
  \shortstack[c]{%
    \makebox[3.2em][r]{$#1$}\\
    $[\makebox[3.2em][r]{$#2$},\makebox[3.2em][r]{$#3$}]$%
  }%
}

\begin{minipage}{0.82\linewidth}
\raggedright
\textbf{(a) Gemma3}~\citep{gemma2025technical}

\vspace{0.25em}

\begin{tabular}{@{}lcc@{}}
\toprule
\textbf{Comparison}
& \shortstack{\textbf{$\Delta$Top-1}\\{[95\% CI]}}
& \shortstack{\textbf{$\Delta$F1}\\{[95\% CI]}} \\
\midrule

ZS $\rightarrow$ ZS+\textsc{SIAB}
& \bootciecc{+9.32}{+7.84}{+10.61}
& \bootciecc{+7.67}{+6.33}{+8.98} \\

Mid $\rightarrow$ Mid+\textsc{SIAB}
& \bootciecc{+3.42}{+2.01}{+4.79}
& \bootciecc{+4.64}{+3.27}{+5.91} \\

High $\rightarrow$ High+\textsc{SIAB}
& \bootciecc{+3.03}{+1.71}{+4.48}
& \bootciecc{+4.25}{+2.83}{+5.54} \\

\textsc{PASTA-ECC} vs.\ ZS
& \bootciecc{+9.14}{+7.80}{+10.60}
& \bootciecc{+8.11}{+6.82}{+9.37} \\

\textsc{PASTA-GoE} vs.\ ZS
& \bootciecc{+1.82}{+0.32}{+3.14}
& \bootciecc{+0.73}{-0.73}{+2.29} \\

\textsc{PASTA}-$\cap$ vs.\ ZS
& \bootciecc{+8.83}{+7.43}{+10.23}
& \bootciecc{+7.76}{+6.37}{+9.21} \\

\textsc{PASTA+SIAB} vs.\ ZS
& \bootciecc{+13.84}{+12.18}{+15.31}
& \bootciecc{+12.51}{+11.23}{+13.84} \\

\bottomrule
\end{tabular}
\end{minipage}
\vspace{0.6em}

\begin{minipage}{0.82\linewidth}
\raggedright
\textbf{(b) Qwen3}~\citep{yang2025qwen3}

\vspace{0.25em}

\begin{tabular}{@{}lcc@{}}
\toprule
\textbf{Comparison}
& \shortstack{\textbf{$\Delta$Top-1}\\{[95\% CI]}}
& \shortstack{\textbf{$\Delta$F1}\\{[95\% CI]}} \\
\midrule

ZS $\rightarrow$ ZS+\textsc{SIAB}
& \bootciecc{+3.30}{+2.10}{+4.46}
& \bootciecc{+2.79}{+1.66}{+3.90} \\
Mid $\rightarrow$ Mid+\textsc{SIAB}
& \bootciecc{+2.08}{+0.94}{+3.20}
& \bootciecc{+1.89}{+0.78}{+2.98} \\
High $\rightarrow$ High+\textsc{SIAB}
& \bootciecc{+1.65}{+0.54}{+2.82}
& \bootciecc{+1.21}{+0.07}{+2.34} \\
\textsc{PASTA-ECC} vs.\ ZS
& \bootciecc{+6.92}{+5.72}{+8.25}
& \bootciecc{+4.60}{+3.46}{+5.76} \\
\textsc{PASTA-GoE} vs.\ ZS
& \bootciecc{+0.09}{-1.00}{+1.21}
& \bootciecc{-0.11}{-1.25}{+0.99} \\
\textsc{PASTA}-$\cap$ vs.\ ZS
& \bootciecc{+4.92}{+3.74}{+6.14}
& \bootciecc{+3.86}{+2.62}{+4.97} \\
\textsc{PASTA+SIAB} vs.\ ZS
& \bootciecc{+10.03}{+8.78}{+11.25}
& \bootciecc{+7.84}{+6.75}{+9.09} \\

\bottomrule
\end{tabular}
\end{minipage}
\vspace{0.6em}

\begin{minipage}{0.82\linewidth}
\raggedright
\textbf{(c) LLaMA3}~\citep{grattafiori2024llama3}

\vspace{0.25em}

\begin{tabular}{@{}lcc@{}}
\toprule
\textbf{Comparison}
& \shortstack{\textbf{$\Delta$Top-1}\\{[95\% CI]}}
& \shortstack{\textbf{$\Delta$F1}\\{[95\% CI]}} \\
\midrule

ZS $\rightarrow$ ZS+\textsc{SIAB}
& \bootciecc{+2.53}{+1.52}{+3.66}
& \bootciecc{+2.95}{+1.94}{+3.91} \\
Mid $\rightarrow$ Mid+\textsc{SIAB}
& \bootciecc{+1.97}{+0.90}{+3.08}
& \bootciecc{+1.66}{+0.63}{+2.71} \\
High $\rightarrow$ High+\textsc{SIAB}
& \bootciecc{+1.25}{+0.19}{+2.32}
& \bootciecc{+1.32}{+0.33}{+2.28} \\
\textsc{PASTA-ECC} vs.\ ZS
& \bootciecc{+6.01}{+4.99}{+7.22}
& \bootciecc{+3.07}{+2.11}{+4.14} \\
\textsc{PASTA-GoE} vs.\ ZS
& \bootciecc{+2.82}{+1.78}{+3.99}
& \bootciecc{+1.46}{+0.32}{+2.45} \\
\textsc{PASTA}-$\cap$ vs.\ ZS
& \bootciecc{+4.01}{+2.96}{+5.03}
& \bootciecc{+0.54}{-0.52}{+1.64} \\
\textsc{PASTA+SIAB} vs.\ ZS
& \bootciecc{+6.21}{+5.18}{+7.40}
& \bootciecc{+3.27}{+2.17}{+4.24} \\

\bottomrule
\end{tabular}
\end{minipage}
\end{table}

\section{Prompt}

\label{app:prompts}
\paragraph{Zero-shot.}
Inputs of the ECC benchmark are rendered as a multi-turn chat (user / assistant
alternation along dialogue turns, assistant role assigned to the
target speaker) via
\texttt{tokenizer.apply\_chat\_template}. A final user message asks
for the emotion of the last utterance with the 10-class label list.
The inputs of the GoEmotions benchmark use a single user message containing the sentence
and the 27-class label list. System message is the default
``You are a helpful assistant'' for both. The ECC final-turn
instruction is:

\begin{promptbox}
What emotion is expressed in the last utterance?
Answer with one word: happy, excited, sad, proud, frustrated,
angry, relaxed, afraid, confident, or guilty.
\end{promptbox}

\paragraph{Few-shot.}
Demonstrations follow an \emph{utterance / cues / reasoning /
emotion} chain-of-thought format, with one demonstration per emotion
class (10 for ECC and 27 for GoEmotions). Demonstrations are prepended
to the ZS prompt and held fixed across methods within each
(dataset, shot) setting. One ECC demonstration is shown below as a template:

\begin{promptbox}
Utterance: "I can't believe I finally got the promotion after
all these years!"
Cues: "can't believe", "finally", "got the promotion"
   -> excitement from achieving a long-awaited goal
Emotion: excited
\end{promptbox}

\section{Bootstrap}
\label{app:bootstrap}
We use paired bootstrap resampling to quantify uncertainty arising from the finite evaluation sets. For each contrast, we draw $B=10{,}000$ samples with replacement from the corresponding test set and recompute each metric for both methods using identical resampled example indices. Model outputs, decoding settings, and hyperparameters remain fixed across replicates. For the GoEmotions benchmark, the temperature $T$ and threshold $t$ are tuned on the development set under the corresponding baseline and then frozen for its $+\textsc{SIAB}$ counterpart. We report the 2.5th and 97.5th percentiles of each paired-difference distribution as a pointwise 95\% confidence interval. SIAB comparisons are prompt-matched, whereas the PASTA-family variants are compared with ZS.

Table~\ref{tab:bootstrap-goe} reports the GoEmotions results, and Table~\ref{tab:bootstrap-ecc} reports the ECC results. Across the two tables, all 18 prompt-matched the proposed SIAB settings yielded positive differences on both reported metrics. Equivalently, all 36 metric-level confidence intervals excluded zero. All 12 metric-level \textsc{PASTA+SIAB}-versus-ZS intervals were also strictly positive. The smallest positive lower bounds occurred under few-shot prompting, including the LLaMA3 model's Mid / High on GoEmotions macro-F1 and the Qwen3 model's High on ECC macro-F1, but remained above zero. In contrast, several stand-alone PASTA intervals included zero, indicating less consistent gains. The largest degradations occurred when ECC-profiled PASTA heads were transferred to GoEmotions, particularly for the LLaMA3 model, although LLaMA3's in-domain \textsc{PASTA-GoE} configuration also decreased the Top-1 accuracy. Overall, the paired-bootstrap results supported the consistency of SIAB across models, datasets, prompting regimes, and evaluation metrics.

\end{document}